\documentclass[runningheads]{llncs}

\usepackage{graphicx}
\usepackage{amsmath,amssymb,amsthm}
\usepackage{booktabs}

\title{Hierarchical Indexing with Knowledge Enrichment for Multilingual Video Corpus Retrieval}
\titlerunning{Hierarchical Indexing with KG Enrichment}
\author{
Yu Wang\inst{1}\thanks{Corresponding author; Email: yuw235@pitt.edu} \and
Tianhao Tan\inst{2} \and
Yifei Wang\inst{3}
}
\authorrunning{Y. Wang et al.}
\institute{
\inst{1}School of Computing and Information, University of Pittsburgh, PA, USA \\
\email{yuw235@pitt.edu}\\
\inst{2}Wuhan University of Technology\\
\email{tantianhao@whut.edu.cn}\\
\inst{3}Hunan University\\
\email{wangyifei0411@hnu.edu.cn}
}
\begin{document}
\maketitle
\begin{abstract}
Retrieving relevant instructional videos from multilingual medical archives is crucial for answering complex, multi-hop questions across language boundaries. However, existing systems either compress hour-long videos into coarse embeddings or incur prohibitive costs for fine-grained matching. We tackle the Multilingual Video Corpus Retrieval (mVCR) task in the NLPCC-2025 M4IVQA challenge with a multi-stage framework that integrates multilingual semantics, domain terminology, and efficient long-form processing. Video subtitles are divided into semantically coherent chunks, enriched with concise knowledge-graph (KG) facts, and organized into a hierarchical tree whose node embeddings are generated by a language-agnostic multilingual encoder. At query time, the same encoder embeds the input question; a coarse-to-fine tree search prunes irrelevant branches, and only the top-ranked chunks are re-scored by a lightweight large language model (LLM). This design avoids exhaustive cross-encoder scoring while preserving chunk-level precision. Experiments on the mVCR test set demonstrate state-of-the-art performance, and ablation studies confirm the complementary contributions of KG enrichment, hierarchical indexing, and targeted LLM re-ranking. The proposed method offers an accurate and scalable solution for multilingual retrieval in specialized medical video collections.

\end{abstract}
\section{Introduction}
{\setlength{\parskip}{0pt}
Online video has become a primary medium for disseminating information, and medical instructional content is increasingly recognized for conveying complex health topics \cite{li2024towards,zhong2025enhancing,wang2025systematic}. However, their sheer volume and unstructured nature make it challenging to locate specific information. This work addresses the Multilingual Video Corpus Retrieval (mVCR) task in the NLPCC-2025 M4IVQA challenge \cite{10.1007/978-981-97-9443-0_38,li2025overview}. The objective is to retrieve the most relevant untrimmed video from a large, multilingual collection, even when the query language differs from the video's subtitles. Effective retrieval faces three key challenges: achieving robust multilingual semantics, efficiently processing lengthy videos rich in medical terminology, and bridging the gap between concise queries and comprehensive video content \cite{kong2022spvit,ke2025detection,kong2023peeling}.

Existing retrieval strategies, however, fail to meet the specific demands of the mVCR task.
Dual-encoder models \cite{mithun2018learning,dong2019dual,yu2018joint} enable fast lookup through shared query–video embeddings but cannot capture the extensive temporal structure of long videos \cite{Yu_2025_WACV}.  
Multilingual alignment techniques, whether via machine translation \cite{talipov2017cross} or unified embedding spaces \cite{vatex2019}, often provide insufficient coverage of the domain-specific terminology crucial to medical content.  
Sophisticated temporal models \cite{ni2022xclip,zhao2022centerclip,zhao2025contextualbanditsunboundedcontext} are typically designed for monolingual data and, therefore, miss essential cross-language semantics.  
Neural re-rankers based on cross-encoders \cite{nogueira2019passage,kong2022spvit} or LLMs \cite{liu2023ranking} deliver precise relevance scores but incur prohibitive computational costs when applied as first-stage filters for large corpora, limiting practical scalability.  
Consequently, no single approach simultaneously achieves efficient multilingual retrieval, fidelity to medical terminology \cite{li2024distinct}, and tractable processing of long videos.

To overcome these limitations, we introduce a multi-stage framework that delivers efficient and accurate mVCR. Subtitles are segmented into semantic chunks, enriched with KG facts, and then organized using LaBSE \cite{LaBSE_paper} embeddings into a hierarchical index for coarse-to-fine retrieval. An embedded query initiates a tree search that prunes irrelevant branches early to reduce search costs. Finally, only the top candidate chunks are re-ranked by a lightweight multilingual LLM, yielding nuanced relevance scores without processing the entire corpus.

The main contributions of this work are:
\par
(1) A modular architecture that combines semantic chunking, domain-specific KG enrichment, and multilingual embeddings to enable scalable search in long medical videos. \par
(2)  A dynamic tree-pruning strategy that balances efficiency and precision by narrowing the search space before LLM re-ranking. \par
(3) State-of-the-art results on the mVCR test set, supported by ablation studies that isolate the impact of KG enrichment, hierarchical indexing, and LLM re-ranking.

\section{Related Work}

Early video retrieval struggled with semantic gaps due to its reliance on low-level features and metadata. Deep dual encoders \cite{mithun2018learning,dong2019dual,yu2018joint} improved scalability via independent indexing of joint query-video embeddings. However, they often neglected fine-grained cross-modal interactions and struggled to represent long temporal sequences \cite{Yu_2025_WACV}. Transformers have further enhanced temporal reasoning, with models like CLIP4Clip~\cite{luo2022clip4clip} demonstrating the efficacy of frame aggregation. Subsequent refinements in temporal alignment using sliding windows or segment attention \cite{ni2022xclip,zhao2022centerclip} also showed promise. Nevertheless, these methods are predominantly monolingual and general-domain, making them unsuitable for domain-specific terminology and multilingual queries in mVCR.

Multilingual retrieval initially relied on machine translation, but poor translation quality limited performance, spurring direct alignment techniques that embed sentences into unified semantic spaces. Models like LaBSE \cite{LaBSE_paper} and other language-agnostic embeddings proved effective for multilingual sentence retrieval and on benchmarks like VATEX \cite{vatex2019}, which contains bilingual captions for short videos. Nonetheless, VATEX's short videos differ significantly from the long, procedural, terminology-dense medical videos in mVCR. Adapting these techniques to specialized, lengthy, code-mixed content remains challenging.

Knowledge Graphs (KGs) offer structured relational information to mitigate vocabulary discrepancies between queries and video content. Early strategies involved query expansion and path-based reasoning for enhanced recall \cite{xiong2017explicit}. VideoGraph \cite{rossetto2021videograph} integrates KG entities with shot-level video features for open-domain search, while BioSyn \cite{biosyn2020} uses UMLS to link clinical terms by synthesizing synonyms, improving retrieval in noisy settings. Despite these advances, most KG-augmented approaches focus on text or short video clips. Effectively incorporating KG information into long, multilingual instructional videos while preserving temporal coherence remains unaddressed.

Large Language Models (LLMs) significantly advanced re-ranking by capturing nuanced query-document semantics \cite{shen2024altgen,zhang2025comparative}. BERT-based cross-encoders \cite{nogueira2019passage,kong2022spvit} substantially outperform traditional passage re-ranking methods. Prompting large generative models for zero-shot relevance scoring also shows consistent multilingual improvements \cite{liu2023ranking}. Their computational expense remains a primary drawback, as cross-encoders scale quadratically with input length, making them unsuitable as first-stage filters for large collections of long videos. This computational bottleneck significantly challenges the application of advanced semantic matching to long-form, multilingual video retrieval.
\vspace{-0.2cm}
\section{Methods}
\label{sec:methods}
Our multi-stage framework addresses the mVCR challenge of retrieving relevant videos when the query and subtitle languages differ. At its core, our pipeline enriches subtitles with KG facts and builds a hierarchical index for each video, enabling efficient coarse-to-fine retrieval. The system consists of two main phases: Hierarchical Index Construction (Section \ref{sec:index_construction}) and Retrieval \& Ranking (Sections \ref{sec:query_retrieval}-\ref{sec:reranking_aggregation}). This design effectively scales to long videos while bridging language gaps between queries and video content.
\subsection{Task Formulation}
\label{sec:task_formulation}

Let $\mathcal{V} = \{v_1, \dots, v_N\}$ denote a collection of $N$ medical instructional videos. Each video $v_i$ is associated with a set of subtitles $\mathcal{S}_i$ (in its original language $L_i$) and a collection of relevant KG triples $\mathcal{K}_i$. Let $\mathcal{Q}$ be the space of user queries, where a query $Q \in \mathcal{Q}$ may be expressed in any supported language $L_Q$. The objective is to learn a retrieval function $f: \mathcal{Q} \times \mathcal{V} \to \text{Ranking}$ that generates a ranked list of videos from $\mathcal{V}$ based on their relevance to a given query $Q$. 
\begin{figure}
\centering
\includegraphics[scale=0.23]{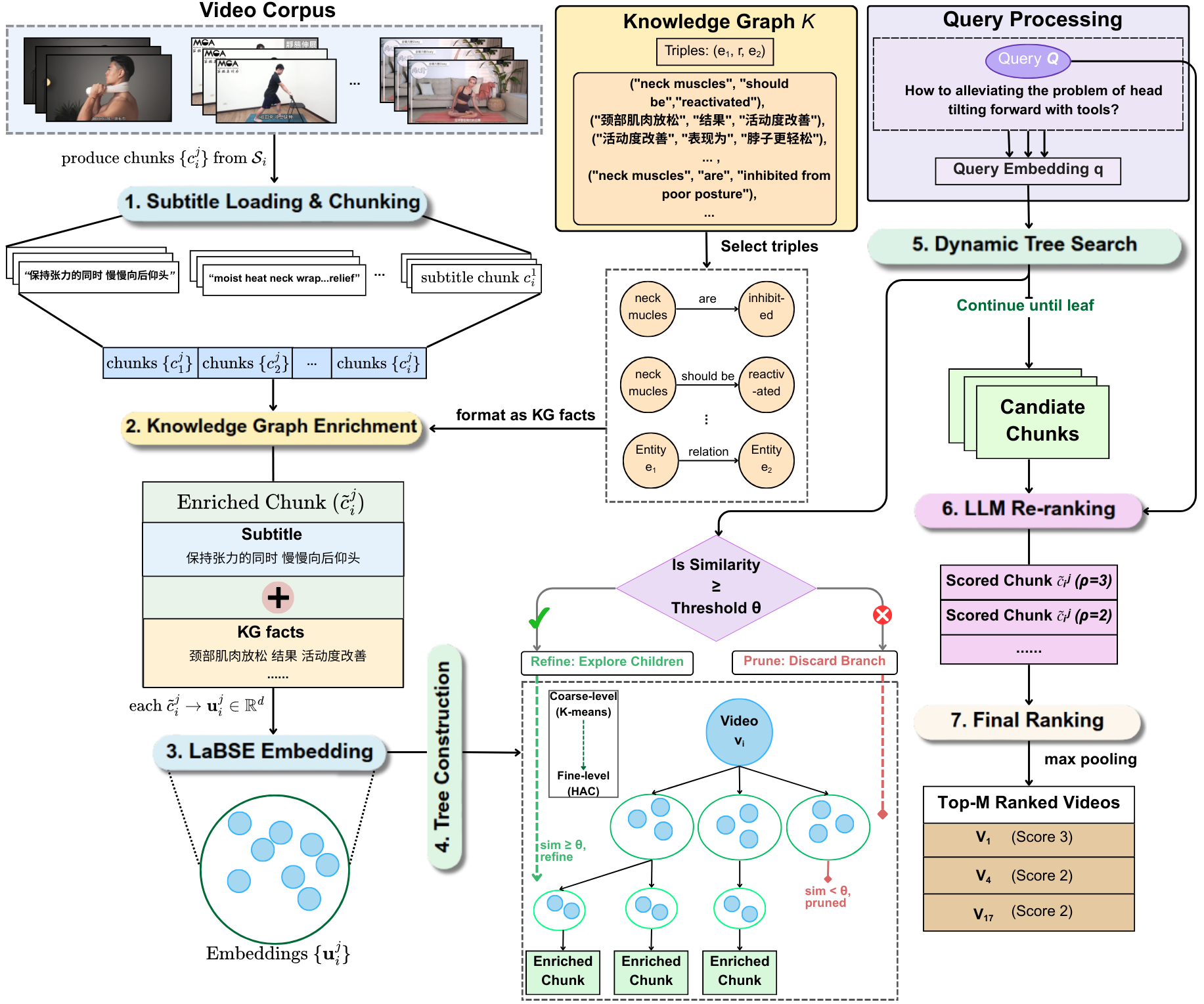}
\caption{Overview of proposed multilingual video retrieval pipeline.} \label{fig1}
\vspace{-0.6cm}
\end{figure}
\subsection{Hierarchical Index Construction}
\label{sec:index_construction}
For each video $v_i \in \mathcal{V}$, we construct an enriched hierarchical index $\mathcal{T}_i$ through three key steps: segmenting subtitles into semantic chunks, enhancing these chunks with relevant KG context, and organizing them into a hierarchical tree structure.\\

\noindent \textbf{Subtitle Loading and Semantic Chunking.} Segmenting lengthy instructional videos requires preserving semantic coherence. We begin by loading subtitle data $\mathcal{S}_i = \{\ell_1, \dots, \ell_{m_i}\}$ for each video and performing basic text cleaning. To avoid disrupting the instructional flow of arbitrary divisions, our approach employs semantic chunking based on the similarity between adjacent subtitle lines. The LaBSE model, with its strong multilingual representation capabilities, provides embeddings for these subtitle lines:
\begin{equation}
\mathbf{e}(\ell_k) = \text{LaBSE}(\ell_k) \in \mathbb{R}^d
\label{eq:line_embedding}
\end{equation}
where $\mathbf{e}(\ell_k)$ represents the $d$-dimensional embedding for subtitle line $\ell_k$. Chunk boundaries are detected by identifying points where the cosine similarity between adjacent embeddings falls below an empirically tuned threshold $\tau$:
\begin{equation}
\cos(\mathbf{e}(\ell_k), \mathbf{e}(\ell_{k+1})) < \tau
\label{eq:chunk_boundary}
\end{equation}
This boundary detection yields a sequence of chunks $c_i^1, \dots, c_i^{M_i}$ for video $v_i$, each containing semantically coherent subtitle lines in the original language $L_i$.\\

\noindent \textbf{KG Enrichment via Text Concatenation.}
To enhance semantic representation with domain knowledge, each chunk $c_i^j$ is enriched using the set of KG triples $\mathcal{K}_i$ relevant to its corresponding video $v_i$. We identify medical entities in $c_i^j$ using a multilingual Named Entity Recognizer, specifically an XLM-RoBERTa model fine-tuned for multilingual NER \cite{XLM_R_paper}. These entities directly guide the selection of relevant knowledge. Specifically, for each identified entity, we retrieve all triples from the video's pre-filtered set $\mathcal{K}_i$ where it appears as either the subject or the object. This enrichment occurs via straightforward text concatenation: key information from the retrieved triples, such as entity relationships and types, is converted into concise, factual statements and appended to the chunk's original text. The resulting enriched chunk, $\tilde{c}_i^j$, thus integrates relevant domain knowledge while preserving the original content structure.\\

\noindent \textbf{Tree Construction.}
The hierarchical organization of enriched chunks forms the foundation of our efficient retrieval approach. We encode each enriched chunk into a vector representation using LaBSE:
\begin{equation}
\mathbf{u}_i^j = \text{LaBSE}(\tilde{c}_i^j)
\label{eq:chunk_embedding}
\end{equation}
where $\mathbf{u}_i^j$ represents the enriched chunk in the shared multilingual semantic space. We implement a two-level clustering strategy to create a structure that captures both broad topics and specific details. First, K-means partitions the set of embeddings $\{\mathbf{u}_i^j\}$ for video $v_i$ into $K$ coarse clusters that serve as first-level nodes representing major topics. Then, for each coarse cluster, we apply HAC to develop deeper tree levels that reveal finer-grained subtopics. Each node $n$ in the resulting tree $\mathcal{T}_i$ stores a representative embedding $\mathbf{e}_n$, calculated as the centroid of $\mathcal{C}_n$, the set of embeddings corresponding to all chunks descended from node $n$:
\begin{equation}
\mathbf{e}_n = \frac{1}{|\mathcal{C}_n|} \sum_{\mathbf{u} \in \mathcal{C}_n} \mathbf{u}
\label{eq:node_embedding}
\end{equation}
This centroid $\mathbf{e}_n$ effectively summarizes the semantic content of node $n$. The completed hierarchical structure $\mathcal{T}_i$, whose leaf nodes are the individual enriched chunks $\tilde{c}_i^j$, enables the efficient coarse-to-fine search.
\vspace{-0.2cm}
\subsection{Query Processing and Initial Retrieval}
\label{sec:query_retrieval}
After constructing hierarchical indices for all videos, our system processes incoming user queries and performs efficient retrieval across these structured representations to identify relevant candidate chunks.\\

\noindent \textbf{Query Processing.}
Given a user query $Q$ in language $L_Q$, we encode it using the same LaBSE model employed during the indexing phase:
\begin{equation}
\mathbf{q} = \text{LaBSE}(Q)
\label{eq:query_embedding}
\end{equation}
This consistent encoding projects the query vector $\mathbf{q}$ and the chunk embeddings $\mathbf{u}_i^j$ into a shared multilingual semantic space, enabling comparisons based on conceptual similarity rather than exact lexical matching. Consequently, the system gains inherent robustness against linguistic variations, such as those introduced by translation. \\

\noindent \textbf{Dynamic Tree Search.}
Rather than exhaustively comparing the query against all chunks, we implement a coarse-to-fine search strategy that leverages the hierarchical index $\mathcal{T}_i$ of each video. The search commences by computing the cosine similarity between the query embedding $\mathbf{q}$ and the embeddings $\mathbf{e}_n$ of the first-level (K-means) cluster nodes. Using a predefined relevance threshold $\theta$, only top-level clusters whose similarity exceeds this threshold are retained, effectively pruning irrelevant branches early in the search. For each qualifying cluster, the search recursively descends its subtree (generated by HAC), applying the same pruning logic at each level to traverse only promising branches. This traversal continues until reaching leaf nodes (the enriched chunks $\tilde{c}_i^j$) via unpruned paths.
The resulting candidate set comprises all leaf nodes reached through this process. Each candidate is recorded as a tuple $(\text{video\_id}, \text{chunk\_id}, \cos(\mathbf{q}, \mathbf{u}_{i}^j), \tilde{c}_{i}^j)$, with its cosine similarity serving as an initial relevance score. These candidates are then ranked to form a top-M list, $\mathcal{C}_\text{top}$, for subsequent re-ranking. This hierarchical search offers significant computational improvements over a brute-force scan. By dynamically pruning the search space, it ensures the expensive LLM re-ranker processes only a few top candidates, which is critical for balancing high precision with the scalability and low latency required for large video corpora.
\vspace{-0.2cm}
\subsection{LLM Re-ranking and Video Aggregation}
\label{sec:reranking_aggregation}

While embedding similarity provides efficient initial retrieval, achieving optimal ranking requires a deeper semantic understanding. In this final phase, we refine the ranking of candidate chunks using a multilingual LLM and then aggregate these scores to produce video-level results.\\

\noindent \textbf{Multilingual LLM Re-ranking.}
Embedding-based similarity captures broad semantic relationships but often misses nuanced relevance factors crucial for medical queries. To address this limitation, we implement LLM-based re-ranking that leverages multilingual understanding capabilities. For each candidate chunk $\tilde{c}_i^j \in \mathcal{C}_\text{top}$, we prompt the LLM with the original query $Q$ (in language $L_Q$) and the chunk's enriched text $\tilde{c}_i^j$ (in its original language $L_i$). The model then produces a scalar relevance rating $\rho(\tilde{c}_i^j)$ on a 1--3 scale (where 3 indicates highest relevance), providing a fine-grained assessment that surpasses simple vector similarity. While the LLM's nuanced semantic understanding leads to highly effective scoring, this performance comes with a trade-off: the model's internal reasoning is largely opaque, limiting its interpretability. \\

\noindent \textbf{Video-level Aggregation and Ranking.}
The mVCR task requires video-level rankings rather than chunk-level results. We aggregate the chunk scores for each video $v_i$ using max pooling, which assigns each video the highest relevance score achieved by any of its evaluated chunks:
\begin{equation}
\text{Score}(v_i) = \max_{\tilde{c}_i^j \in \mathcal{C}_\text{top}^i} \rho(\tilde{c}_i^j)
\label{eq:video_score}
\end{equation}
where $\mathcal{C}_\text{top}^i$ represents the set of top chunks belonging to video $v_i$. This approach prioritizes videos containing highly relevant content chunks, even if other portions are less pertinent. The final output ranks all videos $\{v_1, \dots, v_N\}$ in descending order of $\text{Score}(v_i)$.\\
\vspace{-0.2cm}
\section{Experiments}
\subsection{Dataset and Evaluation}
\label{sec:dataset_eval}

The dataset features medical instructional videos crawled from YouTube, with textual content that includes original subtitles and generated captions in both Chinese and English. The corresponding question-answer pairs consist of Chinese questions manually authored by medical experts and English questions that are translated and refined by native-speaking medical doctors. Each question corresponds to a specific temporal segment of a video, and multiple related questions may point to the same answer segment. The dataset is divided into training, validation, and test sets.
\vspace{-0.3cm}
\begin{table}
\centering
\caption{Statistics of the Medical Instructional Video Dataset, including the number of videos, QA pairs, vocabulary size, and average lengths for the Train/Dev/Test splits.}
\label{tab:dataset_stats}
\begin{tabular}{lccccccc}
\hline
Dataset & Videos & QA pairs & Vocab & Avg. Ch. Q. & Avg. Eng. Q. & Avg. Video \\
 &  &  & Size & Len. & Len. & Len. \\
\hline
Train & 1,228 & 5,840 & 6,582 & 17.16 & 6.97 & 263.3 \\
Dev & 200 & 983 & 1,743 & 17.81 & 7.26 & 242.4 \\
Test & 200 & 1,022 & 2,234 & 18.22 & 7.44 & 310.9 \\
\hline
\end{tabular}
\vspace{-0.3cm}
\end{table}

Following the challenge protocol \cite{pioneering_work_ref}, retrieval performance is measured using Recall@n (R@n) with $n \in \{1, 10, 50\}$, which indicates the percentage of queries where the correct video appears among the top-$n$ results. The Mean Reciprocal Rank (MRR) \cite{IR_metrics_ref} is calculated as:
\begin{equation}
\text{MRR} = \frac{1}{|V|} \sum_{i=1}^{|V|} \frac{1}{\text{Rank}_i}
\end{equation}
where $|V|$ represents the number of test queries, and $\text{Rank}_i$ is the position of the ground-truth video in the predicted list for the $i$-th query. The Overall score, serving as the main ranking criterion, is calculated by summing the four metrics:
\begin{equation}
\text{Overall} = \sum_{i=1}^{|M|} \text{Value}_i
\end{equation}
where $|M|=4$ is the number of evaluation metrics, and $\text{Value}_i$ represents the value of the $i$-th metric.

\subsection{Main Results}
{
\begin{table}
\centering
\vspace{-0.3cm}
\caption{Retrieval performance comparison of our proposed framework against other methods on the mVCR test set.}
\label{tab:results_comparison}
\begin{tabular}{lccccc}
\toprule
Method & R@1 & R@10 & R@50 & MRR & Overall \\
\midrule
RANDOMPICK \cite{pioneering_work_ref,cmivqa_1} & 0.0343 & 0.0523 & 0.0442 & 0.0442 & 0.1674 \\
GEN \cite{NLPCC_Overview_2024} & 0.1311 & 0.1074 & 0.0978 & 0.1142 & 0.4505 \\
Wjh & 0.2744 & 0.3312 & 0.4117 & 0.2551 & 1.2724 \\
DSG-1 \cite{dsg2023,overview_2023} & 0.2644 & 0.3545 & 0.4414 & 0.2887 & 1.3491 \\
sun \cite{sun2024} & 0.3121 & 0.4078 & 0.4966 & 0.3245 & 1.5410 \\
NYU & 0.3213 & 0.4137 & 0.5104 & 0.3354 & 1.5808 \\
DIMA (Ours) & \textbf{0.3264} & \textbf{0.4211} & \textbf{0.5177} & \textbf{0.3407} & \textbf{1.6059} \\
\bottomrule
\end{tabular}
\vspace{-0.3cm}
\end{table}
}
We evaluate our proposed method against several strong baselines from previous mVCR challenges using five standard metrics: R@1, R@10, R@50, MRR, and the Overall score. Key competitors include GEN~\cite{NLPCC_Overview_2024}, which implemented a retrieval framework based on approaches surveyed in recent literature. DSG-1~\cite{dsg2023,overview_2023} developed a two-stage retrieval-reranking pipeline that employed GPT-3.5\footnote[1]{https://poe.com/GPT-3.5-Turbo.} for video summary generation and RoBERTa~\cite{RoBERTa_ref} for initial retrieval, followed by a CCGS-VCR analyzer for re-ranking. The MQuA approach~\cite{sun2024}, from the 2024 challenge, leveraged the DeBERTa-v2-710M-Chinese~\cite{DeBERTa_ref} model combined with Multi-Level Video Moment Refinement (MVMR) and enhanced it with Multilingual Query Paraphrase Generation (MQPG) using Few-shot ChatGPT~\cite{OpenAI2022_ref}. 

As shown in Table~\ref{tab:results_comparison}, our proposed method achieves state-of-the-art performance across all evaluation metrics on the mVCR test set, achieving the highest scores in R@1 (0.3264), R@10 (0.4211), R@50 (0.5177), MRR (0.3407), and Overall (1.6059).

Compared to GEN~\cite{NLPCC_Overview_2024}, our method demonstrates substantial improvements with absolute increases of 0.1953$\uparrow$ in R@1, 0.3137$\uparrow$ in R@10, 0.4199$\uparrow$ in R@50, 0.2265$\uparrow$ in MRR, and 1.1554$\uparrow$ in Overall score. Against the stronger DSG-1 approach~\cite{dsg2023,overview_2023}, we achieve notable improvements: 0.0620$\uparrow$ for R@1, 0.0666$\uparrow$ for R@10, 0.0763$\uparrow$ for R@50, 0.0520$\uparrow$ for MRR, and 0.2568$\uparrow$ for Overall score. Furthermore, our method outperforms the sun team's MQuA approach~\cite{sun2024} with consistent gains of 0.0143$\uparrow$ in R@1, 0.0133$\uparrow$ in R@10, 0.0211$\uparrow$ in R@50, 0.0162$\uparrow$ in MRR, and 0.0649$\uparrow$ in Overall score. \par
These consistent improvements validate the effectiveness of our multi-stage retrieval framework. The performance gains are particularly significant for R@1 and MRR metrics, highlighting the superior precision of our approach in retrieving the most relevant video as the top result for multilingual medical video retrieval tasks.
\vspace{-0.3cm}
\subsection{Ablation Study}
\label{sec:ablation_study}

We conduct ablation experiments to evaluate the impact of each core component by systematically removing modules from our framework. Generally, all reported percentage changes represent relative decreases from the full system's performance. Removing domain-specific KG enrichment decreases the Overall score by 6.6\% and R@1 by 7.6\%. These results confirm KG enrichment's crucial role in enhancing semantic representations, particularly in bridging specialized medical terminology between queries and video content.

When our hierarchical index is replaced with flat retrieval, performance degrades further, with the Overall score dropping by an additional 2.7\%. Notable declines appear in MRR (12.4\%) and R@50 (8.4\%), showing that flat indexing struggles with long-form videos even under relaxed retrieval criteria. The magnitude of this performance gap underscores the effectiveness of the hierarchical organization for efficient pruning and improved precision.

{
\begin{table}
\centering
\vspace{-0.3cm}
\caption{Impact of removing key components from the proposed framework on retrieval performance.}
\label{tab:results_ablation}
\begin{tabular}{lccccc}
\toprule
Method & R@1 & R@10 & R@50 & MRR & Overall \\
\midrule
\textbf{DIMA} & \textbf{0.3264} & \textbf{0.4211} & \textbf{0.5177} & \textbf{0.3407} & \textbf{1.6059} \\
w/o KG Enrichment & 0.3017 & 0.3943 & 0.4878 & 0.3154 & 1.4992 \\
w/o Hierarchical Index & 0.2968 & 0.3887 & 0.4741 & 0.2985 & 1.4581 \\
w/o LLM Re-ranking & 0.2855 & 0.3614 & 0.4612 & 0.2911 & 1.3992 \\
\bottomrule
\end{tabular}
\vspace{-0.3cm}
\end{table}
}
\vspace{-0.1cm}
LLM re-ranking is the most critical component, as evidenced by the substantial 12.9\% decrease in Overall score when removed from the pipeline. Performance metrics show the most significant deterioration here, with R@10 and MRR falling by 14.2\% and 14.6\% respectively. Such pronounced degradation relative to other ablations demonstrates that the LLM's nuanced understanding of query-video semantic relationships is critical for achieving state-of-the-art performance in multilingual medical video retrieval.

\vspace{-0.2cm}

\section{Conclusion}
\vspace{-0.1cm}
We have presented a solution to the mVCR challenge that addresses key limitations in retrieving multilingual medical videos. By integrating language-agnostic embeddings with domain knowledge and efficient hierarchical search, our framework achieves both computational scalability and multilingual precision. Experiments demonstrate state-of-the-art performance across all metrics, and ablation studies confirm that each component makes complementary contributions to the effectiveness of multilingual video retrieval.

To build upon this work, future directions include exploring structured KG reasoning with methods like graph neural networks to overcome the limitations of text concatenation. We will also investigate knowledge distillation to create a more compact and efficient LLM re-ranker. Finally, incorporating visual features for true multi-modal retrieval remains a key priority to further boost precision and scalability.

\vspace{-0.3cm}

%
%
\bibliographystyle{unsrt}
\bibliography{refs}

\begin{thebibliography}{10}

\bibitem{li2024towards}
Shutao Li, Bin Li, Bin Sun, and Yixuan Weng.
\newblock Towards visual-prompt temporal answer grounding in instructional video.
\newblock {\em IEEE Transactions on Pattern Analysis \& Machine Intelligence}, (01):1--18, 2024.

\bibitem{zhong2025enhancing}
Jiachen Zhong and Yiting Wang.
\newblock Enhancing thyroid disease prediction using machine learning: A comparative study of ensemble models and class balancing techniques.
\newblock 2025.

\bibitem{wang2025systematic}
Yiting Wang, Jiachen Zhong, and Rohan Kumar.
\newblock A systematic review of machine learning applications in infectious disease prediction, diagnosis, and outbreak forecasting.
\newblock 2025.

\bibitem{10.1007/978-981-97-9443-0_38}
Bin Li, Yixuan Weng, Qiya Song, Lianhui Liang, Xianwen Min, and Shoujun Zhou.
\newblock Overview of the nlpcc 2024 shared task 7: Multi-lingual medical instructional video question answering.
\newblock In Derek~F. Wong, Zhongyu Wei, and Muyun Yang, editors, {\em Natural Language Processing and Chinese Computing}, pages 429--439, Singapore, 2025. Springer Nature Singapore.

\bibitem{li2025overview}
Bin Li, Shenxi Liu, Yixuan Weng, Yue Du, Yuhang Tian, and Shoujun Zhou.
\newblock Overview of the nlpcc 2025 shared task 4: Multi-modal, multilingual, and multi-hop medical instructional video question answering challenge.
\newblock {\em arXiv preprint arXiv:2505.06814}, 2025.

\bibitem{kong2022spvit}
Zhenglun Kong, Peiyan Dong, Xiaolong Ma, Xin Meng, Wei Niu, Mengshu Sun, Xuan Shen, Geng Yuan, Bin Ren, Hao Tang, et~al.
\newblock Spvit: Enabling faster vision transformers via latency-aware soft token pruning.
\newblock In {\em European conference on computer vision}, pages 620--640. Springer, 2022.

\bibitem{ke2025detection}
Zong Ke, Shicheng Zhou, Yining Zhou, Chia~Hong Chang, and Rong Zhang.
\newblock Detection of ai deepfake and fraud in online payments using gan-based models.
\newblock {\em arXiv preprint arXiv:2501.07033}, 2025.

\bibitem{kong2023peeling}
Zhenglun Kong, Haoyu Ma, Geng Yuan, Mengshu Sun, Yanyue Xie, Peiyan Dong, Xin Meng, Xuan Shen, Hao Tang, Minghai Qin, et~al.
\newblock Peeling the onion: Hierarchical reduction of data redundancy for efficient vision transformer training.
\newblock In {\em Proceedings of the AAAI Conference on Artificial Intelligence}, volume~37, pages 8360--8368, 2023.

\bibitem{mithun2018learning}
Niluthpol~Chowdhury Mithun, Juncheng Li, Florian Metze, and Amit~K. Roy-Chowdhury.
\newblock Learning joint embedding with multimodal cues for cross-modal video-text retrieval.
\newblock In {\em Proceedings of the 2018 ACM on International Conference on Multimedia Retrieval}, ICMR '18, page 19–27, New York, NY, USA, 2018. Association for Computing Machinery.

\bibitem{dong2019dual}
Jianfeng Dong, Xirong Li, Chaoxi Xu, Shouling Ji, Yuan He, Gang Yang, and Xun Wang.
\newblock Dual encoding for zero-example video retrieval, 2019.

\bibitem{yu2018joint}
Youngjae Yu, Jongseok Kim, and Gunhee Kim.
\newblock A joint sequence fusion model for video question answering and retrieval, 2018.

\bibitem{Yu_2025_WACV}
Pinrui Yu, Zhenglun Kong, Pu~Zhao, Peiyan Dong, Hao Tang, Fei Sun, Xue Lin, and Yanzhi Wang.
\newblock Q-tempfusion: Quantization-aware temporal multi-sensor fusion on bird's-eye view representation.
\newblock In {\em Proceedings of the Winter Conference on Applications of Computer Vision (WACV)}, pages 5489--5499, February 2025.

\bibitem{talipov2017cross}
Pavel Braslavski, Suzan Verberne, and Ruslan Talipov.
\newblock Show me how to tie a tie: Evaluation of cross-lingual video retrieval.
\newblock In Norbert Fuhr, Paulo Quaresma, Teresa Gon{\c{c}}alves, Birger Larsen, Krisztian Balog, Craig Macdonald, Linda Cappellato, and Nicola Ferro, editors, {\em Experimental IR Meets Multilinguality, Multimodality, and Interaction}, pages 3--15, Cham, 2016. Springer International Publishing.

\bibitem{vatex2019}
Xin Wang, Jiawei Wu, Junkun Chen, Lei Li, Yuan-Fang Wang, and William~Yang Wang.
\newblock Vatex: A large-scale, high-quality multilingual dataset for video-and-language research, 2020.

\bibitem{ni2022xclip}
Yiwei Ma, Guohai Xu, Xiaoshuai Sun, Ming Yan, Ji~Zhang, and Rongrong Ji.
\newblock X-clip: End-to-end multi-grained contrastive learning for video-text retrieval, 2022.

\bibitem{zhao2022centerclip}
Shuai Zhao, Linchao Zhu, Xiaohan Wang, and Yi~Yang.
\newblock Centerclip: Token clustering for efficient text-video retrieval.
\newblock In {\em Proceedings of the 45th International ACM SIGIR Conference on Research and Development in Information Retrieval}, SIGIR ’22, page 970–981. ACM, July 2022.

\bibitem{zhao2025contextualbanditsunboundedcontext}
Puning Zhao, Rongfei Fan, Shaowei Wang, Li~Shen, Qixin Zhang, Zong Ke, and Tianhang Zheng.
\newblock Contextual bandits for unbounded context distributions, 2025.

\bibitem{nogueira2019passage}
Rodrigo Nogueira and Kyunghyun Cho.
\newblock Passage re-ranking with bert, 2020.

\bibitem{liu2023ranking}
Mofetoluwa Adeyemi, Akintunde Oladipo, Ronak Pradeep, and Jimmy Lin.
\newblock Zero-shot cross-lingual reranking with large language models for low-resource languages, 2023.

\bibitem{li2024distinct}
Bin Li, Bin Sun, Shutao Li, Encheng Chen, Hongru Liu, Yixuan Weng, Yongping Bai, and Meiling Hu.
\newblock Distinct but correct: generating diversified and entity-revised medical response.
\newblock {\em Science China Information Sciences}, 67(3):132106, 2024.

\bibitem{LaBSE_paper}
Fangxiaoyu Feng, Yinfei Yang, Daniel Cer, Naveen Arivazhagan, and Wei Wang.
\newblock Language-agnostic bert sentence embedding, 2022.

\bibitem{luo2022clip4clip}
Huaishao Luo, Lei Ji, Ming Zhong, Yang Chen, Wen Lei, Nan Duan, and Tianrui Li.
\newblock Clip4clip: An empirical study of clip for end to end video clip retrieval, 2021.

\bibitem{xiong2017explicit}
Chenyan Xiong, Russell Power, and Jamie Callan.
\newblock Explicit semantic ranking for academic search via knowledge graph embedding.
\newblock In {\em Proceedings of the 26th International Conference on World Wide Web}, WWW '17, page 1271–1279, Republic and Canton of Geneva, CHE, 2017. International World Wide Web Conferences Steering Committee.

\bibitem{rossetto2021videograph}
Luca Rossetto, Matthias Baumgartner, Narges Ashena, Florian Ruosch, Romana Pernisch, Lucien Heitz, and Abraham Bernstein.
\newblock Videograph – towards using knowledge graphs for interactive video retrieval.
\newblock In {\em MultiMedia Modeling: 27th International Conference, MMM 2021, Prague, Czech Republic, June 22–24, 2021, Proceedings, Part II}, page 417–422, Berlin, Heidelberg, 2021. Springer-Verlag.

\bibitem{biosyn2020}
Mujeen Sung, Hwisang Jeon, Jinhyuk Lee, and Jaewoo Kang.
\newblock Biomedical entity representations with synonym marginalization.
\newblock In Dan Jurafsky, Joyce Chai, Natalie Schluter, and Joel Tetreault, editors, {\em Proceedings of the 58th Annual Meeting of the Association for Computational Linguistics}, pages 3641--3650, Online, July 2020. Association for Computational Linguistics.

\bibitem{shen2024altgen}
Yixian Shen, Hang Zhang, Yanxin Shen, Lun Wang, Chuanqi Shi, Shaoshuai Du, and Yiyi Tao.
\newblock Altgen: Ai-driven alt text generation for enhancing epub accessibility.
\newblock {\em arXiv preprint arXiv:2501.00113}, 2024.

\bibitem{zhang2025comparative}
Hang Zhang, Yanxin Shen, Lun Wang, Chuanqi Shi, Shaoshuai Du, Yiyi Tao, and Yixian Shen.
\newblock Comparative analysis of large language models for context-aware code completion using safim framework.
\newblock {\em arXiv preprint arXiv:2502.15243}, 2025.

\bibitem{XLM_R_paper}
Rahul Mehta and Vasudeva Varma.
\newblock Llm-rm at semeval-2023 task 2: Multilingual complex ner using xlm-roberta.
\newblock In {\em Proceedings of the 17th International Workshop on Semantic Evaluation (SemEval-2023)}, 2023.

\bibitem{pioneering_work_ref}
Bin Li, Yixuan Weng, Bin Sun, and Shutao Li.
\newblock Learning to locate visual answer in video corpus using question.
\newblock In {\em ICASSP 2023 - 2023 IEEE International Conference on Acoustics, Speech and Signal Processing (ICASSP)}, page 1–5. IEEE, June 2023.

\bibitem{IR_metrics_ref}
Olivier Chapelle, Donald Metlzer, Ya~Zhang, and Pierre Grinspan.
\newblock Expected reciprocal rank for graded relevance.
\newblock In {\em Proceedings of the 18th ACM Conference on Information and Knowledge Management}, CIKM '09, page 621–630, New York, NY, USA, 2009. Association for Computing Machinery.

\bibitem{cmivqa_1}
Yixuan Weng and Bin Li.
\newblock Visual answer localization with cross-modal mutual knowledge transfer.
\newblock In {\em ICASSP 2023-2023 IEEE International Conference on Acoustics, Speech and Signal Processing (ICASSP)}, pages 1--5. IEEE, 2023.

\bibitem{NLPCC_Overview_2024}
Bin Li, Yixuan Weng, Qiya Song, Lianhui Liang, Xianwen Min, and Shoujun Zhou.
\newblock Overview of the nlpcc 2024 shared task 7: Multi-lingual medical instructional video question answering.
\newblock In Derek~F. Wong, Zhongyu Wei, and Muyun Yang, editors, {\em Natural Language Processing and Chinese Computing}, pages 429--439, Singapore, 2025. Springer Nature Singapore.

\bibitem{dsg2023}
Ningjie Lei, Jinxiang Cai, Yixin Qian, Zhilong Zheng, Chao Han, Zhiyue Liu, and Qingbao Huang.
\newblock A two-stage chinese medical video retrieval framework with llm.
\newblock In Fei Liu, Nan Duan, Qingting Xu, and Yu~Hong, editors, {\em Natural Language Processing and Chinese Computing}, pages 211--220, Cham, 2023. Springer Nature Switzerland.

\bibitem{overview_2023}
Bin Li, Yixuan Weng, Hu~Guo, Bin Sun, Shutao Li, Yuhao Luo, Mengyao Qi, Xufei Liu, Yuwei Han, Haiwen Liang, Shuting Gao, and Chen Chen.
\newblock Overview of the nlpcc 2023 shared task: Chinese medical instructional video question answering.
\newblock In Fei Liu, Nan Duan, Qingting Xu, and Yu~Hong, editors, {\em Natural Language Processing and Chinese Computing}, pages 233--242, Cham, 2023. Springer Nature Switzerland.

\bibitem{sun2024}
Guyang Yu, Xiaoyang Bi, Jielong Tang, Ming Gu, Tianbai Chen, Zhiqiang Li, and Miankuan Zhu.
\newblock Mqua: Multi-level query-video augmentation for multilingual video corpus retrieval.
\newblock In Derek~F. Wong, Zhongyu Wei, and Muyun Yang, editors, {\em Natural Language Processing and Chinese Computing}, pages 353--364, Singapore, 2025. Springer Nature Singapore.

\bibitem{RoBERTa_ref}
Yinhan Liu, Myle Ott, Naman Goyal, Jingfei Du, Mandar Joshi, Danqi Chen, Omer Levy, Mike Lewis, Luke Zettlemoyer, and Veselin Stoyanov.
\newblock Roberta: A robustly optimized bert pretraining approach, 2019.

\bibitem{DeBERTa_ref}
Pengcheng He, Xiaodong Liu, Jianfeng Gao, and Weizhu Chen.
\newblock Deberta: Decoding-enhanced bert with disentangled attention, 2021.

\bibitem{OpenAI2022_ref}
{OpenAI}.
\newblock Introducing chatgpt: Optimizing language models for dialogue.
\newblock \url{https://openai.com/index/chatgpt/}, November 2022.
\newblock Accessed 04 May 2025.

\end{thebibliography}

\end{document}